\documentclass[lettersize,journal]{IEEEtran}
\usepackage{graphicx} 
\usepackage{amsmath, amssymb, amsthm}

\usepackage[capitalize]{cleveref}
\usepackage{algorithm}
\usepackage{algorithmicx}
\usepackage{algpseudocode}
\usepackage{booktabs}

\usepackage[dvipsnames]{xcolor}

\begin{document}
\title{SPOT: Sensing-augmented Trajectory Planning via Obstacle Threat Modeling}


\author{Chi~Zhang$^{\ast}$,
        Xian~Huang$^{\ast}$,
        and~Wei~Dong%
        \thanks{$^{\ast}$These authors contributed equally to this work.}%
        \thanks{Chi Zhang is with the Robotics, University of Michigan, Ann Arbor, USA. (email: zhc@umich.edu)}%
        \thanks{Xian Huang and Wei Dong are with the School of Mechanical Engineering, Shanghai Jiao Tong University, Shanghai, China. (email: hx2020@sjtu.edu.cn; dr.dongwei@sjtu.edu.cn)}%
        \thanks{Corresponding author: Wei Dong.}%
        }


\maketitle

\begin{abstract}
UAVs equipped with a single depth camera encounter significant challenges in dynamic obstacle avoidance due to limited field of view and inevitable blind spots. While active vision strategies that steer onboard cameras have been proposed to expand sensing coverage, most existing methods separate motion planning from sensing considerations, resulting in less effective and delayed obstacle response.
To address this limitation, we introduce SPOT (Sensing-augmented Planning via Obstacle Threat modeling), a unified planning framework for observation-aware trajectory planning that explicitly incorporates sensing objectives into motion optimization. At the core of our method is a Gaussian Process-based obstacle belief map, which establishes a unified probabilistic representation of both recognized (previously observed) and potential obstacles. This belief is further processed through a collision-aware inference mechanism that transforms spatial uncertainty and trajectory proximity into a time-varying observation urgency map.
By integrating urgency values within the current field of view, we define differentiable objectives that enable real-time, observation-aware trajectory planning with computation times under 10 ms. Simulation and real-world experiments in dynamic, cluttered, and occluded environments show that our method detects potential dynamic obstacles 2.8 seconds earlier than baseline approaches, increasing dynamic obstacle visibility by over 500\%, and enabling safe navigation through cluttered, occluded environments.
\end{abstract}

\IEEEpeerreviewmaketitle

\section{Introduction}

Obstacle avoidance is a fundamental capability for Unmanned Aerial Vehicles (UAVs), particularly in dynamic and unknown environments where prior knowledge is limited or unavailable. Successful navigation in such scenarios critically depends on the UAV’s ability to perceive and reason about its surroundings. Among various onboard sensing technologies, LiDAR (Light Detection and Ranging) provides high mapping accuracy and reliability but remains costly and heavy for lightweight UAV platforms. Depth cameras, in contrast, offer a cost-effective and compact alternative, yet their limited field of view (FoV) severely restricts environmental awareness, especially in cluttered or occluded environments.

To alleviate FoV constraints, several strategies have been explored. Multi-camera systems \cite{Wang2024ICRA} improve spatial coverage at the expense of increased payload and degraded flight agility. Other works exploit UAV yaw rotation to expand coverage using a single camera \cite{Bena2023RAL}, but this approach introduces motion blur, consumes additional energy, and can destabilize flight control. A more flexible alternative is \emph{active vision}, in which the camera orientation is adjusted independently of the UAV body. We adopt this strategy through an Active Vision System (AVS) that enables real-time viewpoint control without disturbing flight dynamics. However, unlike LiDAR, the AVS can only observe one direction at a time. Effective obstacle avoidance therefore requires not only reactive collision avoidance but also proactive reasoning about which regions should be sensed next.

Existing active-vision studies in UAV navigation address this challenge only partially. Some works emphasize object tracking or heuristic view selection~\cite{Lin2024ROBIO}, which perform well in structured scenes with predictable dynamics but lack a principled way to prioritize among multiple simultaneous threats or to reason about occluded, unseen regions. Other approaches focus on local sense-and-avoid strategies~\cite{Chen2021TMECH, Bena2023RAL}, steering sensors using geometrical risk fields or hand-crafted rules; these methods improve reaction to visible hazards but do not maintain a unified probabilistic belief over both observed and unobserved space, preventing coherent optimization of active vision in dynamic, partially observed environments.. This limitation motivates the use of a unified probabilistic obstacle model that can consider both observed and unobserved dynamic obstacles within the same decision process, enabling principled estimation of observation urgency.



\begin{figure}[!t]
    \centering
    \includegraphics[width=0.48\textwidth]{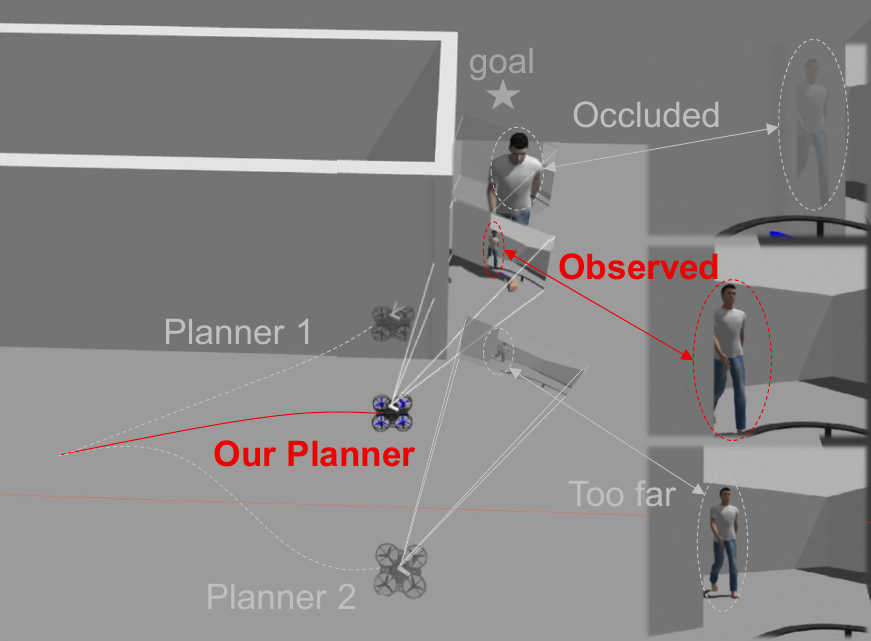}
    \caption{Comparison between independent planner and coupled planner.}
    \label{fig:illustration}
\end{figure}

Even with such a probabilistic model, the UAV must actively adjust its trajectory to reveal occluded threats. When the UAV trajectory is planned without sensing consideration, the vehicle often flies close to obstacles, leading to severe occlusions within the FoV (Fig.~\ref{fig:illustration}). This scenario highlights the importance of sensing-augmented trajectory planning, which balances path efficiency with visibility. Sensing-augmented planners that jointly optimize motion and viewpoint have demonstrated clear benefits in maintaining visibility and reducing perception failures~\cite{Tordesillas2022ACCESS, Tordesillas2023RAL, Wu2022ACCESS, Zhao2022RS}. However, most such frameworks evaluate deterministic visibility or image-quality metrics for already detected objects or prioritize mapping/localization gains for SLAM, and therefore do not explicitly quantify the probability of hidden hazards in blind regions. Therefore, a key challenge remains: how to integrate probabilistic perception and trajectory planning to enable the UAV to both avoid obstacles and actively seek informative viewpoints under uncertainty.

To address these challenges, we propose \textbf{SPOT} (\textbf{S}ensing-augmented \textbf{P}lanning via \textbf{O}bstacle \textbf{T}hreat Modeling), a unified framework that integrates probabilistic obstacle modeling with sensing-augmented trajectory optimization. SPOT constructs a Gaussian-Process (GP)-based obstacle belief that represents both recognized and potential obstacles. This belief representation enables collision risk assessment, which transforms obstacle uncertainty into a time-varying observation urgency map. This urgency map quantifies which regions should be prioritized for sensing based on obstacle uncertainty and proximity to the planned trajectory. Building upon this representation, SPOT formulates an observation-aware trajectory optimization problem. A fast gradient computation scheme based on diffeomorphic integration ensures real-time performance. This unified formulation enables the UAV to plan trajectories that are both safe and perception-effective, proactively steering toward informative viewpoints to maintain robust situational awareness in dynamic and cluttered environments.

The main contributions of this paper are summarized as follows:
\begin{itemize}
    \item We propose SPOT, a unified framework that enables UAV obstacle-aware navigation in dynamic, cluttered, and partially occluded environments.
    \item We construct a Gaussian-Process-based obstacle model that jointly represents both observed and unobserved dynamic obstacles, and derive a collision-aware observation urgency to guide perception-driven planning.
    \item We formulate an observation-aware trajectory planning strategy in which the UAV motion is optimized not only for collision avoidance but also to maximize future perception effectiveness.
\end{itemize}

\section{Related Work}

Obstacle-aware navigation in dynamic and partially observed environments remains a central challenge in UAV planning. In these scenarios, the UAV’s ability to perceive and reason about its surroundings is a key factor for reliable navigation \cite{Wang2021IROS, Zhou2021RAL, Lu2025TRO}. To overcome the limited FoV of a single camera, hardware and platform solutions extend the FoV by using multiple cameras or by rotating the platform. Multi-camera SLAM and multi-FoV systems increase coverage and improve localization accuracy~\cite{Wang2024ICRA, Wang2024ICIRS}, while platform yawing can temporarily enlarge the visible region~\cite{Salaris2011IROS, Zhang2025TAES}. These approaches are effective in increasing raw observability but incur tangible costs in payload, calibration complexity, energy, and flight agility, which limit their suitability for lightweight UAVs operating in cluttered, dynamic scenes.

Active-vision systems \cite{Chen2021TMECH} instead adapt sensing at runtime to improve safety. A large body of work studies information-driven sensing planning for exploration~\cite{Bircher2016ICRA, Kusnur2021ICRA, Spaan2015AAMS, Arora2015ICRA}, visual localization~\cite{Zhang2022RAL, Hua2024TAES} and mapping~\cite{Schmid2020RAL, Meera2019ICRA}. However, they are not designed for obstacle-aware navigation, which requires sensing strategies that actively ensure collision safety.
In response, active-vision approaches for obstacle avoidance have been developed to steer the camera toward high-risk regions using risk-density fields or heuristic rules, and active sense-and-avoid architectures have been demonstrated on flying platforms~\cite{Chen2021TMECH, Bena2023RAL, Lin2024ROBIO, Chen2024TMECH}. While these methods improve detection and reaction in many scenarios, they
often rely on hand-designed or geometry-based measures and lack a unified probabilistic belief over both observed and unobserved dynamic obstacles, limiting principled comparison of observation priorities across different types of threats.


Perception-aware (sensing-augmented) trajectory planners explicitly consider sensing objectives so the vehicle can trade off path efficiency and visibility. Recent works show that jointly optimizing translation and camera yaw, or learning perception-aware controllers, yields substantial gains in maintaining obstacle visibility and reducing perception failures compared to decoupled baselines~\cite{Tordesillas2022ACCESS, Tordesillas2023RAL, Wu2022ACCESS, Zhao2022RS}. However, these frameworks mostly evaluate deterministic visibility, blur, or image-quality metrics for already detected or tracked obstacles, or prioritize mapping/localization gains for SLAM tasks; they generally do not quantify the probability of hazards that may lie in occluded or unobserved regions. As a result, although perception-aware planners improve robustness for observed entities, their deterministic objectives limit anticipatory behavior when threats emerge from blind spots.

Probabilistic spatial modeling, especially Gaussian Process (GP) based representations, provides a principled way to quantify spatial uncertainty and to guide informative sensing and path selection~\cite{Yu2018ICCAS, Kjaergaard2011ICMLA, Popovic2020ICRA, Meera2019ICRA}. GPs have been used for terrain mapping, uncertainty-aware field estimation, and even dynamic obstacle prediction~\cite{Olcay2024ECC, Patterson2019CDC}. These works highlight the value of continuous uncertainty fields for planning, but most treat estimation and planning as separate modules or focus on exploration/information objectives; comparatively few approaches construct a unified probabilistic belief that jointly represents both observed and unobserved dynamic obstacles and embed that belief directly into a real-time, differentiable trajectory optimizer for observation-aware collision avoidance.

In summary, prior research has advanced sensor coverage, active sensing, and perception-aware planning, but the literature still lacks a practical framework that (i) represents obstacle uncertainty over both observed and unobserved dynamic obstacles in a unified, continuous belief, and (ii) converts that belief into an observation urgency metric that is directly embedded in the trajectory optimization objective. This paper addresses these gaps by proposing SPOT, which leverages a GP-based obstacle belief to infer observation urgency and integrates it into sensing-augmented trajectory optimization for real-time UAV navigation under occlusion and uncertainty.

\section{Methodology}
    \begin{figure*}[!t]
    \centering
    \includegraphics[width=\textwidth]{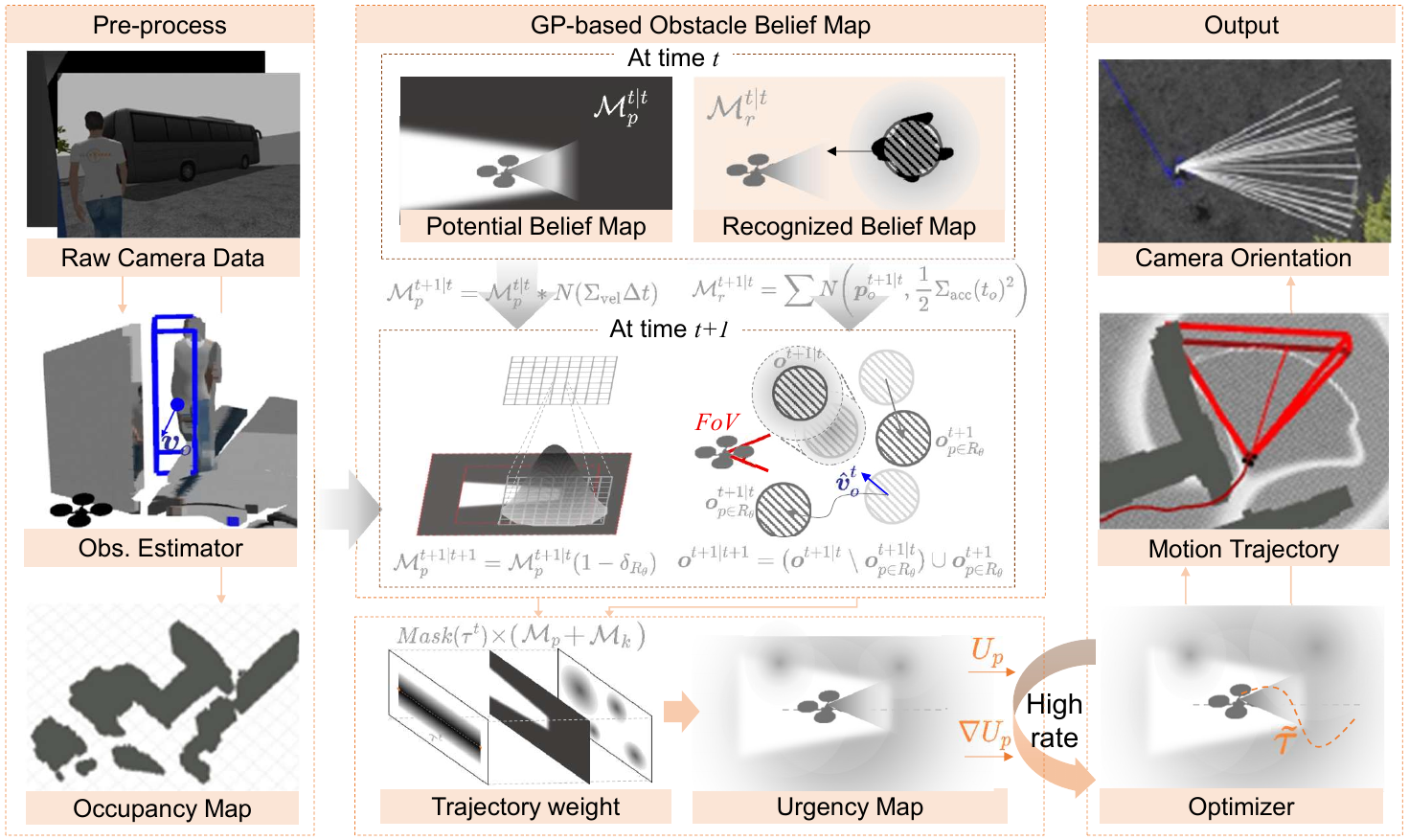}
    \caption{Overview of the SPOT pipeline.}
    \label{fig:pipeline}
\end{figure*}

\subsection{Framework Overview}
We briefly summarize the pipeline of SPOT before presenting each component in detail.
SPOT takes as input a stream of depth observations and UAV states.
It maintains a Gaussian Process (GP)-based obstacle belief map that represents both observed and unobserved regions.
From this belief, an observation urgency field is derived, which encodes where additional sensing would most reduce future collision risk.
This field is then incorporated into a trajectory optimization problem that jointly considers collision avoidance and sensing effectiveness.
The resulting optimization is solved in real time using a gradient-based solver with efficient gradient computation via diffeomorphic integration (Fig.~\ref{fig:pipeline}).

Figure~\ref{fig:pipeline} summarizes the SPOT framework.
Depth observations are first converted into a Gaussian-Process-based obstacle belief map that represents both observed and unobserved regions.
From this belief, an observation urgency field is inferred to quantify which areas require immediate sensing.
Finally, the UAV trajectory and camera orientation are optimized jointly, balancing collision avoidance and perception effectiveness through a real-time gradient-based solver.

\subsection{Obstacle Belief Model}
    We propose a GP-based obstacle modeling framework that estimates a time-evolving probabilistic belief map over both recognized (previously observed) and potential (unobserved) obstacles. This model enables the UAV to balance between reacting to known threats and proactively exploring regions where unseen obstacles may emerge. The resulting belief serves as the foundation for observation urgency estimation, which quantifies collision risk and guides both camera orientation and UAV motion trajectory in an observation-aware planning process.
    
    For computational efficiency and planning relevance, we model the environment as a 2D ground-plane projection of the UAV's 3D operational space. Obstacle states, belief maps, and urgency evaluations are defined in this horizontal plane ($x-y$ space). The UAV’s full 3D trajectory $\boldsymbol{\tau}(t)$ is projected to $\boldsymbol{\tau}_{xy}(t)$ for collision risk assessment and observation modeling.
    
    The belief map is divided into two layers:
    
    \begin{itemize}
        \item $\mathcal{M}_p$: the belief map of potential obstacles in unobserved regions, and
        \item $\mathcal{M}_r$: the belief map of recognized dynamic obstacles, that is, the dynamic obstacles once identified from depth images.
    \end{itemize}
    
    The evolution of $\mathcal{M}_p$ and $\mathcal{M}_r$ is governed by two interacting models: the obstacle motion model, which captures positional uncertainty growth due to motion, and the sensor model, which reflects information gain through AVS-based active perception.
    
    
    To simplify notation, we denote the probability density function of a 2-D Gaussian distribution $\mathcal{N}(\mu, \Sigma)$ with mean $\mu$ and variance $\Sigma$ as
    \begin{equation}
        N \left( x; \mu, \Sigma \right) := \frac{1}{\sqrt{ \left( 2\pi \right)^2 | \Sigma |}} e^{ -\frac{1}{2} (x - \mu)^T \Sigma^{-1} (x - \mu) }
    \end{equation}
    
    Let $\boldsymbol{o} = \{ \boldsymbol{p}_o, \boldsymbol{v}_o, t_o \}$ denote the state of a recognized dynamic obstacle, where $\boldsymbol{p}_o$ is position, $\boldsymbol{v}_o$ is velocity, and $t_o$ is the elapsed time since last observation.
    
    We follow the notation convention from Bayesian filtering. The superscript notation $^{t+1|t}$ denotes a predicted value at time $t+1$ given all information available up to time $t$, while $^{t|t}$ represents an updated (posterior) estimate at time $t$ incorporating observations from the same time step. Additionally, notations such as $^{t+1}$ without conditioning (e.g., $\boldsymbol{o}^{t+1}$) refer to raw sensor observations at that time step.

    \subsubsection{Obstacle Motion Model}
        The obstacle motion model accounts for the positional uncertainty of both potential and recognized dynamic obstacles using stochastic dynamics modeled by Gaussian distributions. Initially, potential obstacles are uniformly distributed over the entire map, and no recognized dynamic obstacles are present:
        \begin{equation}
            \mathcal{M}_p^{0|0}=P_{prior}, \quad \mathcal{M}_r^{0|0}=0
        \end{equation}
        
        Potential obstacle motion is modeled with a zero-mean 2D Gaussian velocity distribution with covariance $\Sigma_{\text{vel}}$. This captures the assumption that potential obstacles move randomly in unknown directions, inducing spatial uncertainty. Recognized dynamic obstacles retain a recorded velocity $\boldsymbol{v}_o$, while their acceleration follows a zero-mean 2D Gaussian with covariance $\Sigma_{\text{acc}}$. These assumptions directly lead to the update rules for the belief map as follows.
        
        The belief map for potential obstacles evolves by convolving the current map with the Gaussian motion kernel:
        \begin{equation}
            \mathcal{M}_p^{t+1|t}=\mathcal{M}_p^{t|t} \ast N(\cdot; 0, \Sigma_{\text{vel}} \Delta t)
        \end{equation}
        
        This convolution captures the spatial diffusion of uncertainty due to   motion variance, reflecting the uncertainty induced by the Gaussian distribution of velocity.
        
        Recognized dynamic obstacles are propagated using their velocities, and the time since last observation $t_o$ is incremented:
        \begin{equation}
            \boldsymbol{p}_o^{t+1|t}=\boldsymbol{p}_o^{t|t} + \boldsymbol{v}_o\Delta t, \quad t_o^{t+1|t}=t_o^{t|t}+\Delta t
        \end{equation}
        
        The belief map of recognized dynamic obstacles is then:
        \begin{equation}
            \mathcal{M}_r^{t+1|t} = \sum N \left( \cdot; \boldsymbol{p}_o^{t+1|t}, \frac{1}{2} \Sigma_{\text{acc}} (t_o^{t+1|t})^2 \right)
        \end{equation}
        
        The variance term $\frac{1}{2} \Sigma_{\text{acc}}(t_o)^2$ is derived from integrating acceleration noise over time, leading to quadratic uncertainty growth. This propagation reflects how position uncertainty accumulates due to unknown acceleration modeled as a zero-mean Gaussian.
    
    \subsubsection{Sensor Model}
        The sensor model describes how observations reduce uncertainty. The belief of potential obstacles in $R_\theta$ is suppressed:
        \begin{equation}
            \mathcal{M}_p^{t+1|t+1} = \mathcal{M}_p^{t+1|t} ( 1 - \delta_{R_\theta} )
        \end{equation}
        
        Here, $\delta_{R_\theta}$ is an indicator function equal to 1 in the field of view and 0 elsewhere. This step clears out potential threats from newly observed regions, reflecting the information gain from active sensing.
        
        Simultaneously, when an obstacle $\boldsymbol{o} = \{ \boldsymbol{p}_o, \boldsymbol{v}_o, t_o \}$ is observed within the current field of view $R_\theta$, its state is updated using the latest measurements:
        \begin{equation}
            \boldsymbol{o}^{t+1|t+1} = (\boldsymbol{o}^{t+1|t} \setminus \boldsymbol{o}_{p \in R_\theta}^{t+1|t}) \cup \boldsymbol{o}_{p \in R_\theta}^{t+1}, \quad \boldsymbol{o} = \{ \boldsymbol{p}_o, \boldsymbol{v}_o, t_o \}
        \end{equation}
        
        Here, previously estimated obstacle states in $R_\theta$ are replaced by new sensor observations, effectively resetting their uncertainty and observation time.
        
        The updated belief map for recognized dynamic obstacles becomes:
        \begin{equation}
            \mathcal{M}_r^{t+1|t+1} = \sum N \left( \cdot; \boldsymbol{p}_o^{t+1|t+1}, \frac{1}{2} \Sigma_{\text{acc}} (t_o^{t+1|t+1})^2 \right)
        \end{equation}
        
        \begin{algorithm}[H]
            \caption{Recursive Obstacle Belief Update via GP-based Prediction and Observation}
            \begin{algorithmic}[1]
                \State \textbf{Input:} $\mathcal{M}_p^{t|t}, \mathcal{M}_r^{t|t}, R_\theta, \boldsymbol{\tau}(t)$ \Comment{Belief maps, FOV, motion trajectory}
                \State \textbf{Initialize:} $\mathcal{M}_p^{0|0} = P_{prior}, \mathcal{M}_r^{0|0} = 0$
                \For{each time step $t$}
                \State $\mathcal{M}_p^{t+1|t} = \mathcal{M}_p^{t|t} \ast f_{\mathcal{N}}(0, \sigma_{vel} \Delta t)$
                \State $\boldsymbol{p}_o^{t+1|t} = \boldsymbol{p}_o^{t|t} + \boldsymbol{v}_o \Delta t, \quad t_o^{t+1|t}=t_o^{t|t}+\Delta t$
                \If{observation available in $R_\theta$}
                \State Mask $\mathcal{M}_p^{t+1|t+1}$ with $1 - \delta_{R_\theta}$
                \State Replace $\boldsymbol{p}_o$, $\boldsymbol{v}_o$ with sensor values
                \State Reset $t_o$
                \EndIf
                \EndFor
            \end{algorithmic}
        \end{algorithm}

\subsection{Collision-aware Observation Urgency}
    To quantify the need for further observation, we define \textit{observation urgency} as a cumulative collision risk metric between predicted obstacles and the UAV's projected motion trajectory $\boldsymbol{\tau}_{xy}$ in the horizontal plane. This measure reflects the need for additional sensing in regions with high positional uncertainty and close proximity to the UAV path. Intuitively, obstacles that are both likely to exist and lie near the UAV trajectory are prioritized for observation.

    \subsubsection{Urgency for Potential Obstacles}
        For a planar spatial location $\boldsymbol{p}$ where a potential (unobserved) obstacle may exist, the urgency at time $t_0$ is defined as:
        
        \begin{align}
            U_p(\boldsymbol{p}, \boldsymbol{\tau}) = \int_{t \in [0,T]} \lambda_p(t) \mathcal{M}_p(\boldsymbol{p}) \cdot \phi_p \left( \boldsymbol{p} \right) \cdot J_{\text{vel}}^{\boldsymbol{\tau}, \boldsymbol{p}}(t) dt
        \end{align}
        
        This formulation integral accumulates weighted collision likelihoods across time. The velocity required for a potential obstacle at location $\boldsymbol{p}$ to reach motion trajectory $\boldsymbol{\tau}_{xy}$ at future time $t$ is $\boldsymbol{v}(t) = \frac{\boldsymbol{\tau}_{xy}(t_0 + t) - \boldsymbol{p}}{t}$, whose probability under the zero-mean Gaussian prior becomes:  
        \begin{align}
            N \left( \boldsymbol{v}(t); 0, \Sigma_{\text{vel}} \right) = N \left( \boldsymbol{p}; \boldsymbol{\tau}_{xy}(t_0 + t), \Sigma_{\text{vel}} t \right) := \phi_p \left( \boldsymbol{p} \right)
        \end{align}
        
        $\mathcal{M}_p(\boldsymbol{p})$ is the probability density of a potential obstacle at position $\boldsymbol{p}$. The function $\lambda_p(t)$ is a non-negative, time-dependent weighting function that enables temporal discounting within the planning horizon.
        
        To ensure consistency with the velocity-based probability space, we introduce a Jacobian term:
        \begin{align}
             J_{\text{vel}}^{\boldsymbol{\tau}, \boldsymbol{p}}(t) := \frac{\partial }{\partial t} \left\| \frac{\boldsymbol{\tau}_{xy}(t_0+t) - \boldsymbol{p}}{t} \right\| _2
        \end{align}
        
        This term accounts for how the spatial offset between $\boldsymbol{p}$ and the motion trajectory maps into velocity space, ensuring probabilistic correctness under the Gaussian motion prior.
    
    \subsubsection{Urgency for Recognized Dynamic Obstacles}
        For each recognized dynamic obstacle $o$ with estimated position $\boldsymbol{p}_o^{t_0}$ and last observed velocity $\boldsymbol{v}_o$, the urgency at time $t_0$ is defined as:

        \begin{align}
            U_r^o (\boldsymbol{\tau}) = \int_{t \in [0,T]} \lambda_r(t) \cdot \phi_r \left( \boldsymbol{p}_o^{t_0 + t} \right) \cdot J_{\text{acc}}^{\boldsymbol{\tau}, \boldsymbol{o}}(t) dt
        \end{align}
        
        This formulation reflects the time-weighted integral accumulation collision probability due to acceleration-driven uncertainty. The Jacobian adjusts for transformation into acceleration space under the Gaussian acceleration prior.
        
        Similarly, for a recognized dynamic obstacle $o$, the required relative acceleration for it to reach motion trajectory $\boldsymbol{\tau}_{xy}$ at future time $t$ is:
        \begin{align}
            \boldsymbol{a}(t) = \frac{2[\boldsymbol{\tau}_{xy}(t_0 + t) - \boldsymbol{p}_o^{t_0 + t}]}{(t - t_o)^2}
        \end{align}
        
        Under the Gaussian acceleration prior, this results in a probability density:
        \begin{align}
            N \left( \boldsymbol{a}(t); 0, \Sigma_{\text{acc}} \right) & = N \left( \boldsymbol{p}_o^{t_0 + t}; \boldsymbol{\tau}_{xy}(t_0 + t), \tfrac{1}{2} \Sigma_{\text{acc}} (t - t_o)^2 \right) \notag \\
            & := \phi_r \left( \boldsymbol{p}_o^{t_0 + t} \right)
        \end{align}
        
        Here, $t_o$ is the time elapsed since the last observation of the obstacle. The function $\lambda_r(t)$ is a non-negative, time-dependent weighting function that enables temporal discounting within the planning horizon. The predicted obstacle position evolves as:
        \begin{equation}
            \boldsymbol{p}_o^{t_0 + t} = \boldsymbol{p}_o^{t_0} + \boldsymbol{v}_o t
        \end{equation}
        
        To ensure consistency with the acceleration-based probability space, we define the Jacobian term:
        \begin{align}
            J_{\text{acc}}^{\boldsymbol{\tau}, \boldsymbol{o}}(t) := \frac{\partial}{\partial t} \left\| \frac{2[\boldsymbol{\tau}_{xy}(t_0+t) - \boldsymbol{p}_o^{t_0 + t}]}{(t - t_o)^2} \right\|_2
        \end{align}
        
        This ensures the probability is correctly accumulated in the time domain under stochastic acceleration dynamics.
    
    \subsubsection{Region-wise Urgency and Camera Orientation}
        The cumulative urgency over a region $R$ is obtained by aggregating the point-wise urgency of both potential and recognized dynamic obstacles:
        \begin{equation}
            U_p^{R, \boldsymbol{\tau}} = \iint_{\boldsymbol{p} \in R} U_p(\boldsymbol{p}, \boldsymbol{\tau}) \, d\boldsymbol{p}, \quad U_r^{R, \boldsymbol{\tau}} = \sum_{\boldsymbol{p}_o \in R} U_r^o (\boldsymbol{\tau})
        \end{equation}
        
        These region-level scores guide observation-aware planning by prioritizing areas with higher estimated collision risk.
        
        Given a motion trajectory $\boldsymbol{\tau}$, the camera yaw orientation $\theta$ is optimized to maximize the cumulative urgency coverage within the camera’s field of view. The observation region $R_\theta$ is modeled as a sector (fan-shaped region) of radius equal to the camera sensing range and central angle determined by the field-of-view specification. The orientation optimization problem is formulated as:
        \begin{equation}
            \theta^\ast = \operatorname*{arg\, max}_{\theta} \left( U_p^{R_\theta, \boldsymbol{\tau}} + U_r^{R_\theta, \boldsymbol{\tau}} \right)
        \end{equation}
        
        This formulation directs the camera toward the sector with the highest estimated collision risk, thereby allocating sensing resources to the most safety-critical regions.

\subsection{Sensing-augmented Trajectory Planning}
    In traditional decoupled planning frameworks, UAV motion planning and sensing are handled as separate tasks, with trajectories optimized mainly for safety and efficiency, and sensing actions designed independently to improve coverage or information gain. However, such separation overlooks the intrinsic coupling between motion and perception: the trajectory determines the sensor’s viewpoint, while the sensed environment influences the trajectory’s safety and feasibility.

    Building upon the observation urgency model introduced earlier, we propose a sensing-augmented planning approach to optimize the UAV’s motion trajectory $\boldsymbol{\tau}$ by incorporating observation urgency information. This formulation enables the UAV to adapt its motion trajectory to reduce collision risk and improve future observation. By integrating sensing considerations into motion planning, the UAV can reduce environmental uncertainty and enhance navigation robustness.
    
    \subsubsection{Sensing-augmented Optimization}
        The UAV motion trajectory $\boldsymbol{\tau}$, represented as a sequence of discrete control points,
        \begin{equation}
            \boldsymbol{\tau} = \{ \boldsymbol{\tau}_1, \dots, \boldsymbol{\tau}_{N_c} \}, \quad \boldsymbol{\tau}_i \in \mathbb{R}^3,
        \end{equation}
        
        is optimized to minimize a cost function integrating observation, collision, smoothness, and dynamic feasibility terms \cite{Zhou2021RAL}, each weighted by user-defined parameters $\lambda$:
        \begin{equation}
            \mathcal{J} (\boldsymbol{\tau}) = \lambda_v \mathcal{J}_v (\boldsymbol{\tau}) + \lambda_c \mathcal{J}_c (\boldsymbol{\tau}) + \lambda_s \mathcal{J}_s (\boldsymbol{\tau}) + \lambda_d \mathcal{J}_d (\boldsymbol{\tau})
        \end{equation}
        
        The urgency-driven observation cost $\mathcal{J}_v$ integrates sensing considerations into trajectory planning by encouraging motion through regions where future perception improvement is most needed. For trajectory optimization, sensing is modeled as a circular region $R_{\text{circ}}(\boldsymbol{\tau}_{xy,i})$ centered at the projection of $\boldsymbol{\tau}_i$ onto the 2D map with radius equal to the sensing range:
        \begin{equation}
            \mathcal{J}_v (\boldsymbol{\tau}, \theta) = \sum_{i=1}^{N_c} - U_p^{R_{\text{circ}} (\boldsymbol{\tau}_{xy, i}), \boldsymbol{\tau}_i}
        \end{equation}
        
        The collision cost $\mathcal{J}_c$ penalizes proximity to predicted obstacles:
        \begin{equation}
            \mathcal{J}_c (\boldsymbol{\tau}) = \sum_{i=1}^{N_c} j_c (\tau _i) = \sum_{i=1}^{N_c} \sum_{j=1}^{N_p} j_c (i,j)
        \end{equation}
        
        The smoothness cost $\mathcal{J}_s$ penalizes large accelerations and jerks to ensure trajectory continuity:
        \begin{equation}
            \mathcal{J}_s (\boldsymbol{\tau}) = \sum_{i=1}^{N_c - 2} \left\| \mathbf{A}_i \right\|_2^2 + \sum_{i=1}^{N_c - 3} \left\| \mathbf{J}_i \right\|_2^2
        \end{equation}
        
        The dynamic feasibility cost $\mathcal{J}_d$ ensures that kinematic limits on velocity, acceleration, and jerk are respected, using a soft penalty function $F(\cdot)$:
        \begin{equation}
            \mathcal{J}_d (\boldsymbol{\tau}) = \sum_{i=1}^{N_c - 1} \omega_v F(\mathbf{V}_i) + \sum_{i=1}^{N_c - 2} \omega_a F(\mathbf{A}_i) + \sum_{i=1}^{N_c - 3} \omega_j F(\mathbf{J}_i)
        \end{equation}
        
        The optimization problem is:
        \begin{equation}
            \boldsymbol{\tau}^* = \operatorname*{arg\,min}_{\boldsymbol{\tau}} \mathcal{J} (\boldsymbol{\tau})
        \end{equation}
        
        yielding a trajectory that is safe, smooth, and observation-aware.
    
    \subsubsection{Real-time Optimization}
        In the online setting, the trajectory $\boldsymbol{\tau}$ is optimized using the L-BFGS algorithm, which requires gradients of all cost components. The gradients $\nabla_{\boldsymbol{\tau}} \mathcal{J}_c$, $\nabla_{\boldsymbol{\tau}} \mathcal{J}_s$, and $\nabla_{\boldsymbol{\tau}} \mathcal{J}_d$ are computed directly from their definitions. 
        
        For the observation term $\mathcal{J}_v$, the gradient with respect to $\boldsymbol{\tau}_i$ is approximated analytically using the divergence theorem over the circular boundary $\partial R_{\text{circ}}$:
        \begin{align}
            & \frac{\partial U_t^{R_{\text{circ}}(\boldsymbol{\tau}_{xy,i}), \boldsymbol{\tau}_i}}{\partial \tau_i} = \\ & \left[ \oint_{\partial R_{\text{circ}}} U_p(x, y, \boldsymbol{\tau}_i) \, dy, \notag \quad -\oint_{\partial R_{\text{circ}}} U_p(x, y, \boldsymbol{\tau}_i) \, dx \right]
        \end{align}
        
        \begin{equation}
            \nabla_{\boldsymbol{\tau}} \mathcal{J}_v (\boldsymbol{\tau}) = \sum_{i=1}^{N_c} - \frac{\partial U_t^{R_{\text{circ}}(\boldsymbol{\tau}_{xy,i}), \boldsymbol{\tau}_i}}{\partial \tau_i}
        \end{equation}
        
        This formulation enables efficient and differentiable trajectory updates during replanning, improving both environmental awareness and navigational safety.

\section{Result}
    \subsection{Experiment Setup}
    We conducted both simulation and real-world experiments to evaluate the proposed method.
    
    The UAV platform is equipped with an active vision system consisting of an Intel RealSense D435i depth camera and a FEETECH SMS3235 servo motor for camera actuation. An UpBoard Plus onboard computer runs the proposed planning algorithm, while flight control is managed by a Pixhawk 6C autopilot.
    
    In all experiments, grid map generation, dynamic obstacle estimation, and avoidance modules remain identical. The discrete planning time step $\Delta t$ is 0.1 s, and the resolution of the grid map, potential obstacle map, recognized obstacle map, and urgency map is 0.1 m. Additional parameters are listed in Tab.~\ref{tab:parameters}.
    
    Dynamic obstacle estimation uses depth images from the active vision system to generate U-depth maps and 3D bounding boxes. A Kalman filter estimates each obstacle’s position, size, and velocity, providing predictions for path planning and active vision control.

\subsection{Simulation Experiments}
    \begin{figure*}[!t]
        \centering
        \includegraphics[width=1\textwidth]{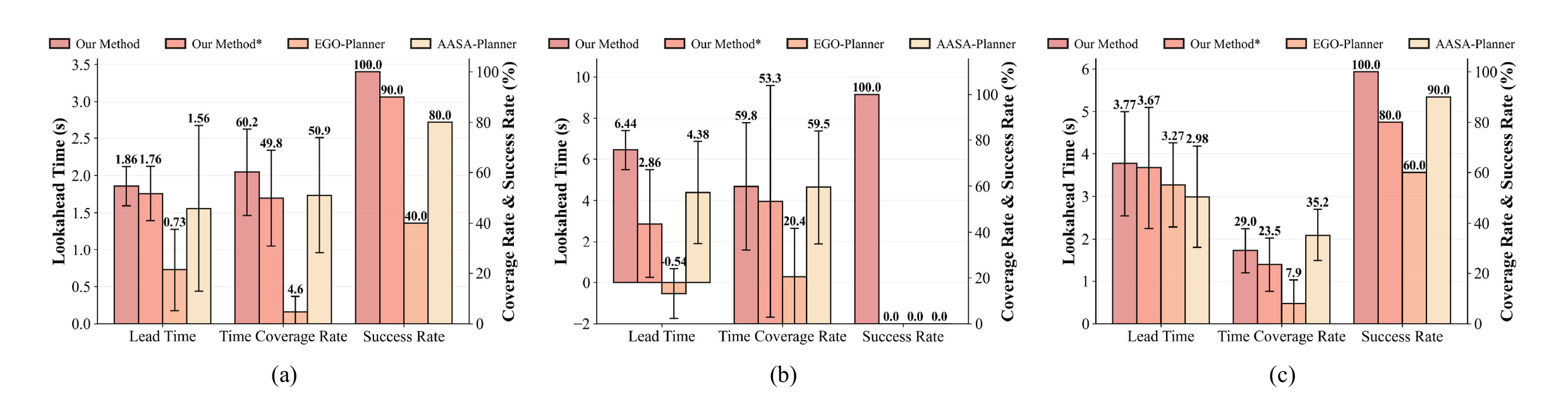}
        \caption{Performance Comparison of Different Planning Methods in Different Scenarios: (a) Street scenario. (b) Simple corner scenario. (c) Museum scenario.}
        \label{fig:sim_bar}
    \end{figure*}
    
    \begin{figure}[htbp]
        \centering
        \includegraphics[width=0.48\textwidth]{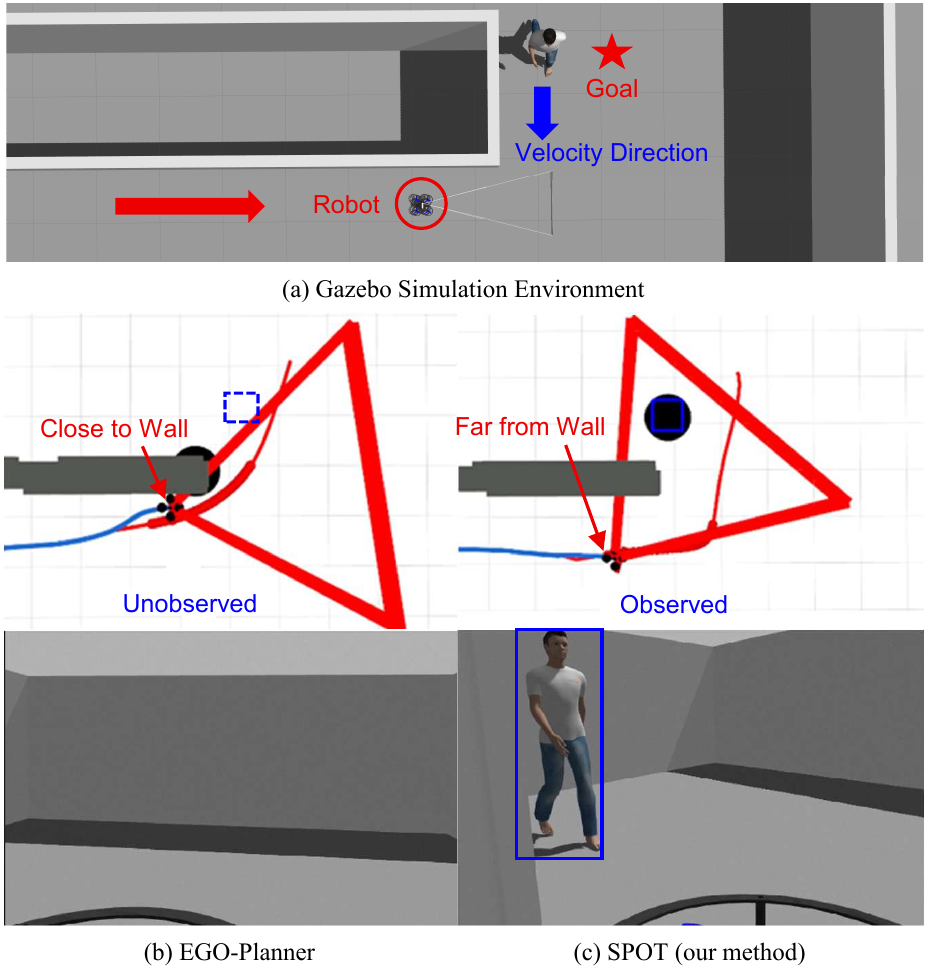}
        \caption{Trajectory comparison in a simple corner environment: (a) Gazebo simulation environment (b) EGO-Planner's trajectory resulting in collision with the obstacle. (c) Our method's trajectory demonstrating successful obstacle avoidance.}
        \label{fig:simple_corner}
    \end{figure}
    
    \subsubsection{Test Scenarios}
        Simulation experiments were performed in the PX4–GAZEBO platform. Three dynamic environments were designed:
        
        \begin{itemize}
            \item \textbf{Street}: an outdoor-like straight path where pedestrians move back and forth;
            \item \textbf{Museum}: an indoor maze-like environment with long corridors and occlusions;
            \item \textbf{Simple corner}: a corner-shaped corridor where dynamic obstacles appear suddenly from behind the wall.
        \end{itemize}
        
        In the street and museum scenarios, each human obstacle moves repeatedly between two fixed points. In the simple corner scenario (Fig.~\ref{fig:simple_corner}), designed to test obstacle avoidance in blind spots, the UAV must navigate around a 90° corner to reach the goal. Once the UAV’s $x$-coordinate exceeds 5m, a simulated human obstacle begins moving forward at a constant velocity from behind the wall, creating a sudden occlusion challenge.

    \subsubsection{Compared Methods}
        We benchmarked four planners:
        \begin{itemize}
            \item \textbf{SPOT}: full sensor-augmented planning with urgency-driven cost;
            \item \textbf{SPOT*}: our method with $\lambda_v = 0$, removing the urgency term;
            \item \textbf{EGO-Planner}: baseline motion planning without sensing objectives;
            \item \textbf{AASA + EGO-Planner}: active sensing with decoupled motion planning.
        \end{itemize}
    
    \subsubsection{Evaluation Metrics}
        Performance was evaluated using three metrics: Success ratio, the percentage of collision-free trials; Observation lead time, the time difference between the first observation of a dynamic obstacle and its entry within distance $d_f$; and Observation time coverage ratio, the percentage of time a dynamic obstacle remains in view while within $d_f$.
    
    \subsubsection{Results}
        Each scenario was tested in 10 trials per method. In the simple corner scenario, the proposed method maintained greater clearance from the wall, increasing observation utility and enabling earlier detection of the approaching obstacle, whereas EGO-Planner remained close to the wall, resulting in late detection and collision. As shown in Tab.~\ref{tab:comparison}, our method achieved the highest success ratio, longest lead time, and best time coverage across all scenarios, confirming its ability to detect and track dynamic obstacles earlier and for longer durations.
        
        \begin{table}[htbp]
          \centering
          \caption{Experiment Parameters}
          \label{tab:parameters}
          \begin{tabular}{ccccccccc}
            \toprule
            $\Delta t$ & $P_{\text{prior}}$ & $\lambda_d$ & $\lambda_v$ & $\lambda_s$ & $\lambda_c$ & $\lambda_r$ & $\sigma_v$ & $\sigma_a$ \\
            \midrule
            0.1 & 0.2 & 0.03 & 0.25 & 1.0 & 0.5 & 6.0 & 2.0 & 10.0\\
            \bottomrule
          \end{tabular}
        \end{table}
        
        \begin{table}[htbp]
          \centering
          \caption{Performance Comparison of Different Planning Methods}
          \label{tab:comparison}
          \begin{tabular}{lccc}
            \toprule
            \textbf{Method} & \textbf{Average Success} & \textbf{Observation} & \textbf{Time Coverage} \\
            & \textbf{Ratio (\%)} & \textbf{Lead Time (s)} & \textbf{Ratio (\%)} \\
            \midrule
            Our Method  & \textbf{100.00} & \textbf{4.02} & \textbf{49.67} \\
            Our Method*  & 64.00 & 2.76 & 41.59 \\
            EGO-Planner  & 40.00 & 1.15 & 10.98 \\
            AASA-Planner & 72.00 & 3.06 & 48.53 \\
            \bottomrule
          \end{tabular}
        \end{table}
        
\subsection{Real-world Experiments}
\begin{figure*}[!t]
        \centering
        \includegraphics[width=1\textwidth]{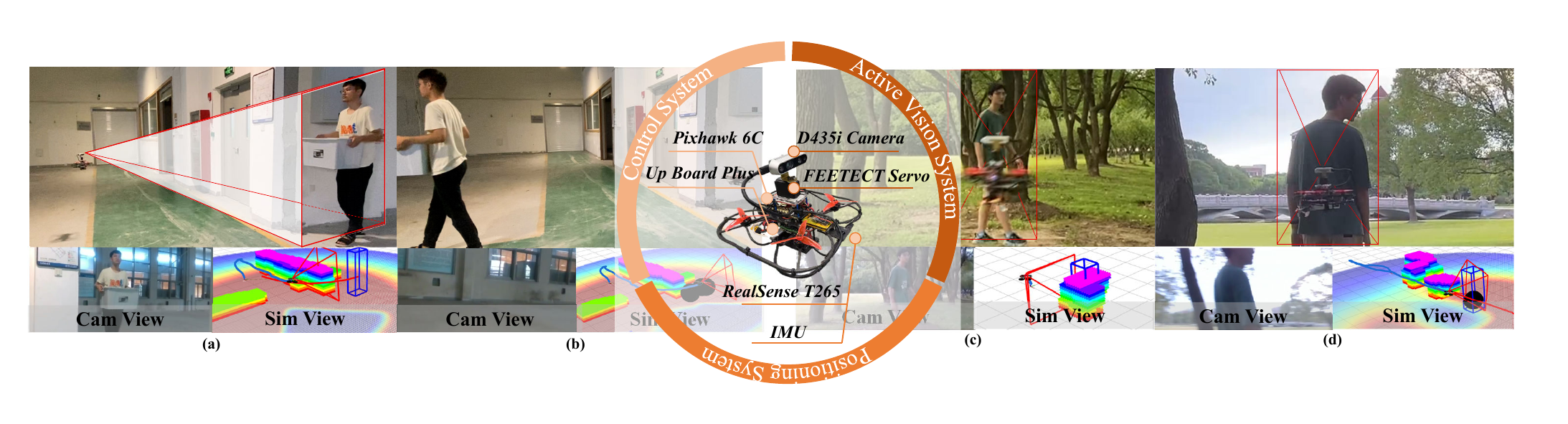}
        \caption{Real-world Experiment and our UAV system. The first row shows the third-person perspective, the second row shows the camera view from the drone, and the simulation map visualization. (a)-(b) depict the indoor corner environment, where the drone avoids suddenly appearing obstacles. (c)-(d) show the outdoor forest environment, where the drone navigates through a forest with multiple dynamic obstacles.}
        \label{fig:drone_and_exp}
    \end{figure*}

    To verify the effectiveness of our algorithm in real-world environments, we conducted two indoor experiments and one outdoor experiments using the UAV shown in Fig.~\ref{fig:drone_and_exp}. Some experiment results are shown in Fig.~\ref{fig:drone_and_exp}. Our system remained consistent across both indoor and outdoor experiments without any modifications. In all these experiments, the UAV utilized T265 for localization data and employed the dynamic obstacle estimator mentioned in the first section to estimate dynamic obstacles.

    In the first indoor experiment, the UAV was tasked with navigating back and forth through a narrow environment containing a dynamic obstacle. Throughout the flight mission, the system demonstrated robust real-time perception capabilities by detecting the dynamic obstacle well in advance of potential collision scenarios. The UAV subsequently executed trajectory planning algorithms to generate safe avoidance paths while maintaining continuous flight operations. Notably, the system exhibited simultaneous multi-target tracking performance, successfully monitoring the position and movement patterns of the dynamic obstacle while concurrently maintaining situational awareness of unknown static obstacles within the environment. This dual-tracking capability enabled the UAV to complete the entire navigation mission without collision incidents. Comprehensive experimental details and flight performance data are documented in the supplementary video material.
    
    The second indoor experiment was designed to simulate a challenging corner navigation scenario, where a dynamic obstacle would suddenly emerge from the blind spot around the corner, as illustrated in the corresponding Fig.~\ref{fig:drone_and_exp}(a)-(b). To address the limited visibility inherent in such corner environments, our UAV implementation employed a proactive exploration strategy, planning more extensive trajectories toward the wall surfaces to maximize observation urgency values and enhance environmental awareness. This strategic approach proved critical when the dynamic obstacle suddenly appeared, as the enhanced observation positioning enabled rapid detection and response. Upon successful identification of the emerging dynamic obstacle, the UAV's real-time planning system immediately engaged, computing and executing an optimized avoidance trajectory that ensured safe passage while maintaining mission objectives. The experimental results demonstrate the effectiveness of the observation urgency-based planning approach for handling sudden dynamic obstacles in geometrically constrained environments.

    \begin{figure}[!t]
        \centering
        \includegraphics[width=0.48\textwidth]{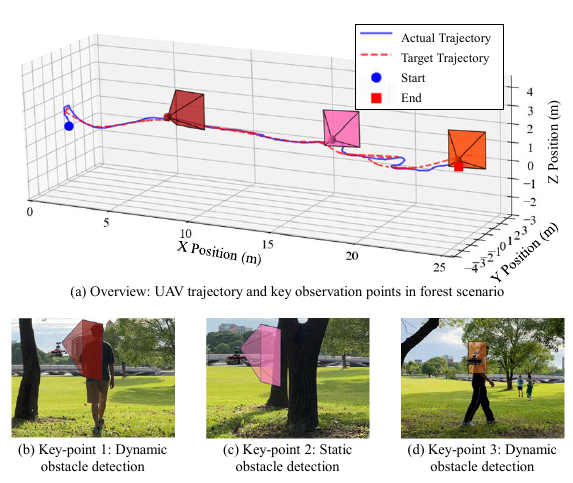}
        \caption{UAV trajectory and obstacle detection in forest scenario: (a) Overview of trajectory with three key observation points, (b-d) Corresponding real-world scenes at each key point. }
        \label{fig:exp_forest}
    \end{figure}

    In the outdoor experiment presented in Fig.~\ref{fig:exp_forest}, we validated the UAV's flight performance in a complex forest environment. The mission required the UAV to navigate through dense wooded terrain while maintaining awareness of potential dynamic obstacles distributed throughout the forest canopy and understory. The system successfully demonstrated the ability to balance comprehensive observation of the intricate static environment with predictive tracking of dynamic obstacles, ensuring stable flight operations and mission completion through the forest traversal task.
    
\begin{figure}[htbp]
        \centering
        \includegraphics[width=0.5\textwidth]{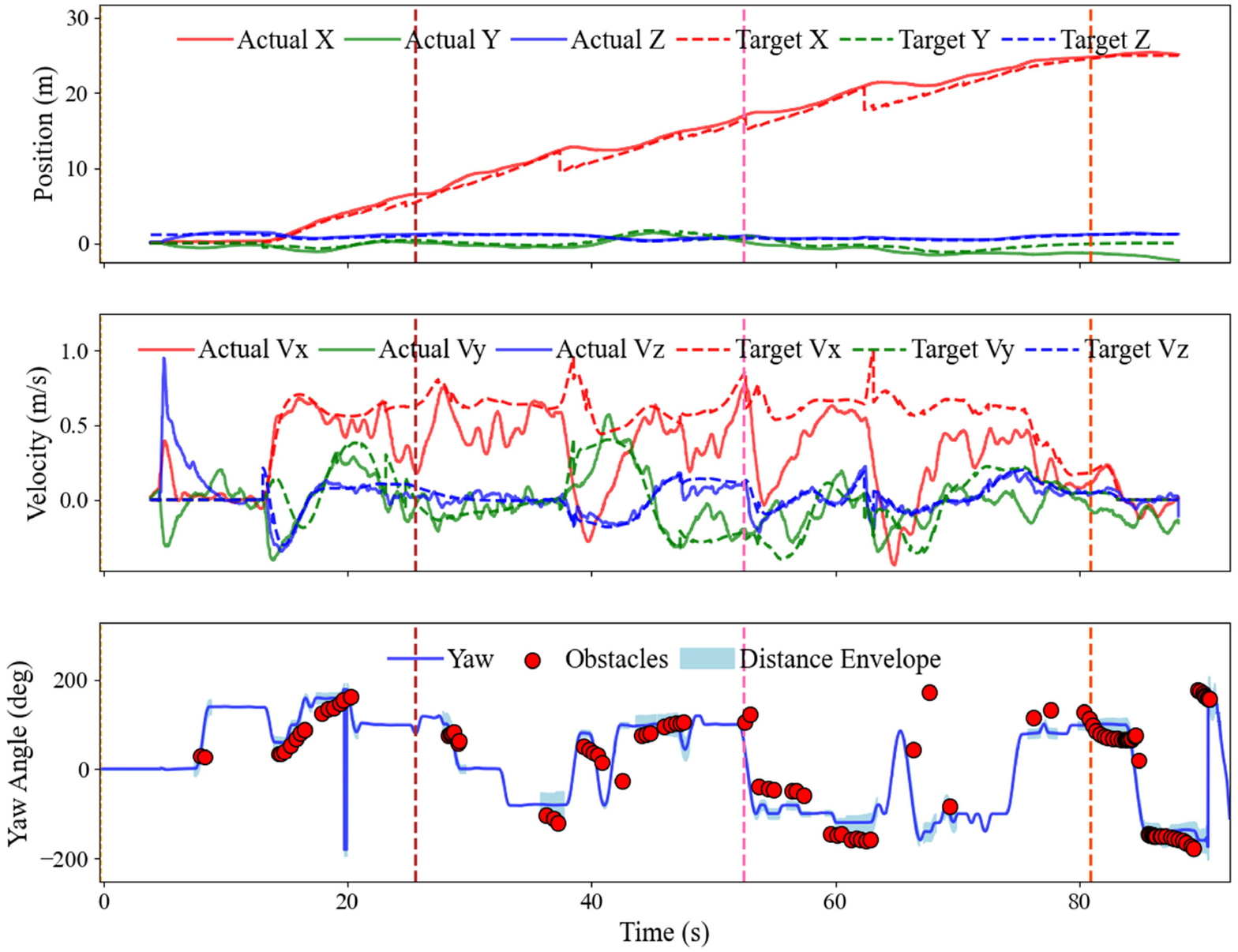}
        \caption{UAV position, velocity, and camera angle data for this experiment. The red, pink, and orange red vertical lines correspond to the time instances of the corresponding colors in Fig.~\ref{fig:exp_forest}. The camera angle plots also show the angles at which dynamic obstacles were observed and relative distance to the UAV.}
        \label{fig:exp_pic_forest_data}
    \end{figure}

    The flight performance data presented in Fig.~\ref{fig:exp_pic_forest_data} provides quantitative validation of the system's capabilities. The upper two plots illustrate the UAV's trajectory tracking performance, comparing actual position against planned position and actual velocity against commanded velocity profiles throughout the mission duration. The third plot reveals the relationship between the UAV's observation angles and the angular positions and distances of detected obstacles, providing insight into the system's perceptual coverage and range capabilities. The experimental data confirms that the UAV successfully detected and tracked multiple dynamic obstacles while effectively balancing observational resources between dynamic obstacle monitoring and static environment mapping, thereby validating the proposed observation urgency framework in real-world forest scenarios.
    
    These three experiments demonstrate that our proposed algorithm can be applied across various environments, whether complex forest settings or narrow indoor spaces. Our algorithm can predict in real-time the positions where potential obstacles are most likely to interfere with UAV flight, enabling observation and avoidance while ensuring safe UAV operation. Throughout the flight process, the UAV's planner was configured to generate plans every 80ms, with threat map generation averaging approximately 10ms and gradient computation taking around 1 ms, thus guaranteeing real-time planning capabilities and flight safety.

\section{Conclusion}
    This paper presents a unified planning framework for UAV navigation in unknown dynamic environments by tightly coupling motion planning with sensor orientation control. We introduce an Active Vision System (AVS) that enhances perception using a single steerable depth camera. A Gaussian Process-based obstacle model jointly captures the uncertainty of both recognized and potential obstacles, enabling the definition of a probabilistically grounded observation urgency metric. This metric guides the UAV to both avoid collisions and improve future observations through strategic motion and sensing. We formulate a coupled optimization problem to minimize collision risk while satisfying smoothness and feasibility constraints, and we implement an efficient real-time solution. Simulation and real-world experiments validate the effectiveness of our approach, demonstrating its superiority over decoupled and heuristic baselines in both safety and awareness. Our work highlights the importance of integrating perception objectives into trajectory planning and offers a scalable framework for UAVs operating in complex, uncertain environments.

In future work, we will extend the GP-based belief model to capture more complex obstacle dynamics and investigate learning-enhanced urgency estimation, as well as explore multi-UAV coordination with distributed active vision systems.

\bibliographystyle{IEEEtran}
\bibliography{refs}





\end{document}